\pdfoutput=1

\documentclass[11pt,a4paper]{article}
\usepackage[final]{emnlp2021}
\usepackage{times}
\usepackage{latexsym}

\usepackage{times}
\usepackage{amsmath}
\usepackage{amssymb} 
\usepackage{latexsym}
\usepackage{graphicx}
\usepackage{booktabs}
\usepackage{tikz}
\usepackage{lipsum}
\usepackage{makecell}
\usepackage{multirow}
\usepackage{array}
\usepackage{bbm}
\usepackage{multicol}
\usepackage{blindtext}
\usepackage{natbib}
\usepackage{caption}
\usepackage{subcaption}
\usepackage{graphicx}

\usepackage{lipsum}
\usepackage{makecell}
\title{Hybrid Rule-Neural Coreference Resolution System based on Actor-Critic Learning (Long Version)}
\author{Yu Wang and Hongxia Jin \\
  AI Center, Samsung Research America \\}

\begin{document}
\maketitle
\begin{abstract}
A coreference resolution system is to cluster all mentions that refer to the same entity in a given context. All coreference resolution systems need to tackle two main tasks: one task is to detect all of the potential mentions, and the other is to learn the linking of an antecedent for each possible mention. 
In this paper, we propose a hybrid rule-neural coreference resolution system based on actor-critic learning, such that it can achieve better coreference performance by leveraging the advantages from both the heuristic rules and a neural conference model. This end-to-end system can also perform both mention detection and resolution by leveraging a joint training algorithm.  We experiment on the BERT model to generate input span representations. Our model with the BERT span representation achieves the state-of-the-art performance among the models on the CoNLL-2012 Shared Task English Test Set.
\end{abstract}

\section{Introduction}
Coreference resolution system contains two main tasks: one task is to detect the valid mentions appearing in a given context, and the other is to group mentions into different clusters such that mentions in each cluster point to the same target entity. The coreference systems can be either heuristic rule-based models \cite{haghighi2009simple,lee2011stanford} or feature-based machine learning structures \cite{bjorkelund2014learning,lee2017end,durrett2013easy,fernandes2012latent,bengtson2008understanding}. One of the main challenges of these systems is that they requires hand-crafted heuristic rules and features, which can be difficult to scale and hardly cover all coreference relations.

A recent trend is the use of neural-based coreference systems, which can avoid most of those hand-engineered features and rules. One of the early works is the end-to-end neural coreference resolution system proposed by \cite{lee2017end}. There are two key scores in a neural coreference model; one is the mention score, and the other is the antecedent score. The two scores accordingly correspond to the mention detection subtask and mention resolution subtask. It is a common practice to calculate these two scores separately using decoupled models. In \cite{zhang2018neural}, however, a joint loss function is leveraged to concurrently optimize the performance of both tasks. Additionally, in the same work, a biaffine attention model is used to compute antecedent scores such that better performance can be obtained in comparison to the models using simple feed-forward networks.

Despite the decent performance exhibited by these neural coreference models, there are still many remaining challenges. One is that there may exist many varieties for a single entity's mentions, and it is very unlikely that our training data can explicitly label all of these singleton mentions. Exiting models suffer from mention proposal \cite{zhang2018neural}. Also most of the neural-based models are effective at identifying and clustering mentions in a document context which are similar to that in the training dataset, but perform much worse once the mention's document context changes. This is primarily because most neural-based models are trained to understand the mention links only using the documents given by the training dataset. Once the document contexts change a lot and are very different from what the systems have seen in the training dataset, the model cannot link the coreferent mentions or entities correctly base on what the system learnt from training data.


To this end, we propose a new {\bf{h}}ybrid {\bf{r}}ule-{\bf{n}}eural {\bf{a}}ctor-{\bf{c}}ritic-based {\bf{c}}oreference system (HRN-ACC), to better handle the mention varieties and document context changes, by leveraging both the heuristic rules and an actor-critic neural coreference system.

This proposed system gives us the advantages that it can significantly alleviate the weaknesses of the rule-based and neural-based coreference systems and keep their advantages at the same time. To be specifically, the new system can cover the coreference relations which cannot be scaled by the rule-based system, and it can also alleviate the negative impact when the document context in test data is different from that in training data, which may leads to wrong coreference clusters for a given mention in different contexts. 

To further improve the neural-based part of system, we leverage an actor-critic deep reinforcement learning network structure. In this structure, we introduce a distance based reward function such that it can take the distance between two mentions into consideration as co-reference is sensitive to the mention distance. The actor-critic model itself can also better handle the mention's stochasticity in the training data (\emph{i.e.} the same mentions appear in different contexts) by leveraging the deep reinforcement learning technique. Finally, the proposed system is an end-to-end system, which can jointly achieve mention detection and mention clustering by using an augmented loss function. Similar to the earlier neural co-reference models giving state-of-the-art performance with BERT models, we also use BERT technique to generate our input embeddings in order to achieve the best performance.

We evaluate the system on the CoNLL-2012 English dataset and achieve the new state-of-the-art performance of 88.6$\%$ average F1 score by using this new model with BERT representations.

\section{System Structure}
In this section, the detailed structure of the new proposed hybrid rule-neural actor-critic-based coreference system (HRN-ACC) is given. First we show how to generate the \emph{mention span representations} in our system, which is inspired by the model given in \cite{lee2017end}.
Here a possible \emph{span} is an N-gram within
a single document. We consider all possible spans
up to a predefined maximum width.
We will also explain the coreference rules used in our hybrid system, and the actor-critic structure for extracting the coreference entities from a document. Technical details are also given on training the hybrid rule-neural model model in semi-supervised fashion by maximizing the expected rewards using the actor-critic deep reinforcement learning model, as well as jointly training mention clustering and mention detection using the augmented actor-critic loss functions.

\subsection{Span Representation}

In this section, we present our span representations by using a bidirectional encoder representations from transformers (BERT) model \cite{devlin2018bert}.
Figure \ref{span2} shows the model structure used to generate span representations $m_i$ ($0 \leq i \leq k$) for all valid mention entities, each mention entity may contain one or several tokens $w_t$. For each token $w_t$ in the sentence, its embedding $o_t$ is generated by the BERT embedding layer. The generated embeddings are then passed to an attention layer with their outputs as the head-finding attention vetctors $w_i^{head}$ for the mention spans:

\begin{figure}[ht!]
  \centering
  \includegraphics[width=1.05\linewidth]{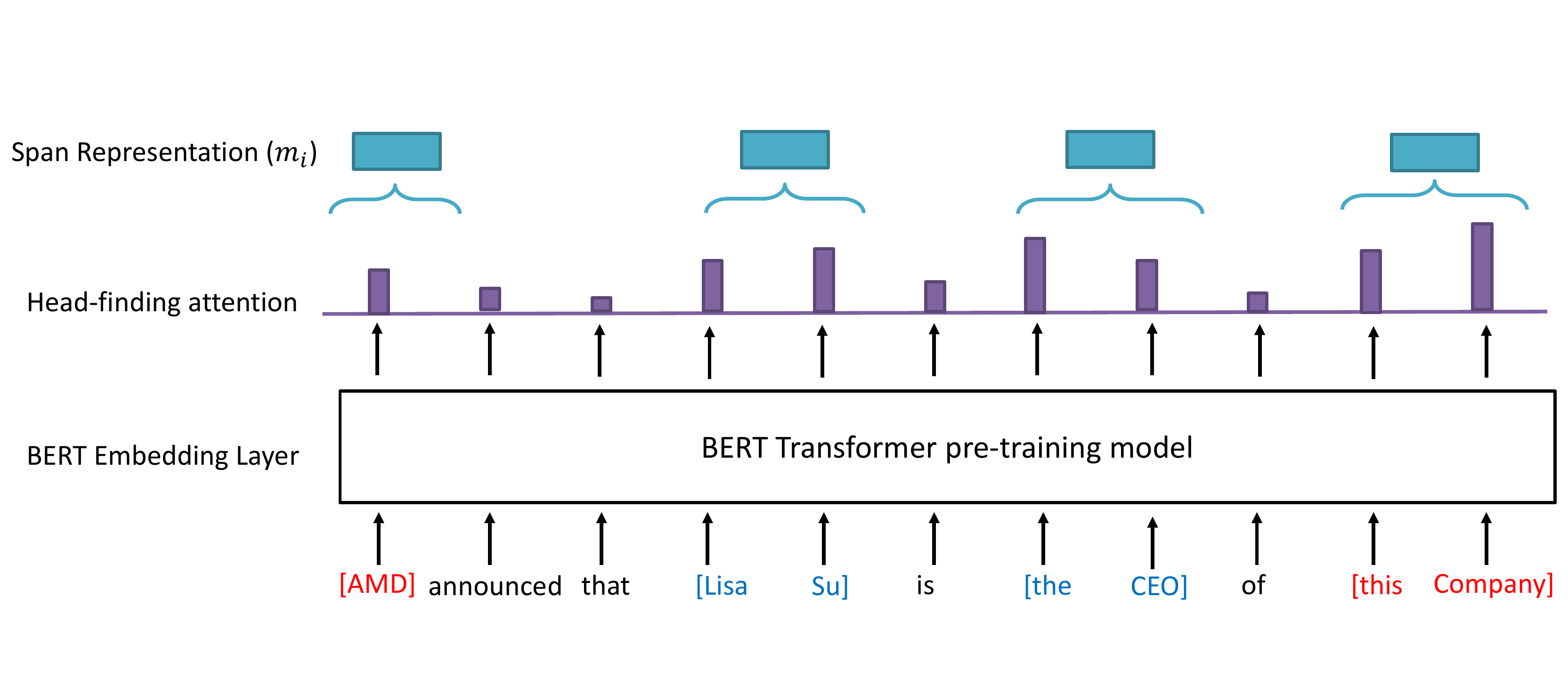}
  \caption{The model structure for generating a BERT Transformer span representations}
  \label{span2}
\end{figure}
\begin{equation}
\begin{split}
o_{t}&={v}_{o}^Tf_{bert}(w_t)\\
{\alpha}_{i,t}&=\dfrac{e^{o_t}}{\sum_{k=i_{start}}^{i_{end}}e^{o_k}}\\
w_i^{head}&=\sum_{k=i_{start}}^{i_{end}}{\alpha}_{i,t}w_t
\end{split}
\label{spanrepresentation}
\end{equation}
where ${\alpha}_{i,t}$ is the word-level attention parameter for the $t^{th}$ word in the $i^{th}$ mention, and $w_i^{head}$ is the head-finding attention vector.

$i_{start}$ stands for the starting word position in the $i^{th}$ mention, and correspondingly $i_{end}$ is the ending word position. 

Inspired by a similar setup as in \cite{zhang2018neural}, the span presentation is a concatenation of four vectors, which is defined as:
\begin{equation}
m_i=o_{i_{start}}\oplus o_{i_{end}}\oplus w_i^{head} \oplus \lambda_{i}
\label{spanrepresentation2}
\end{equation}


\subsubsection{Neural Mentioned Detection Model}
We directly leverage the BERT model's outputs by passing the mention span representations to a feed-forward neural network and a sigmoid function to assign each span a probability score. By ranking all the mentions spans based on there probability scores, the top \emph{K} mentions are selected as our mention candidates. These candidates are then passed into the hybrid rule-neural actor-critic coreference model.

\subsection{Hybrid Rule-Neural Actor-Critic Coreference (HRN-ACC) Model Design}
\begin{figure*}[ht!]
  \centering
  \includegraphics[width=0.83\linewidth]{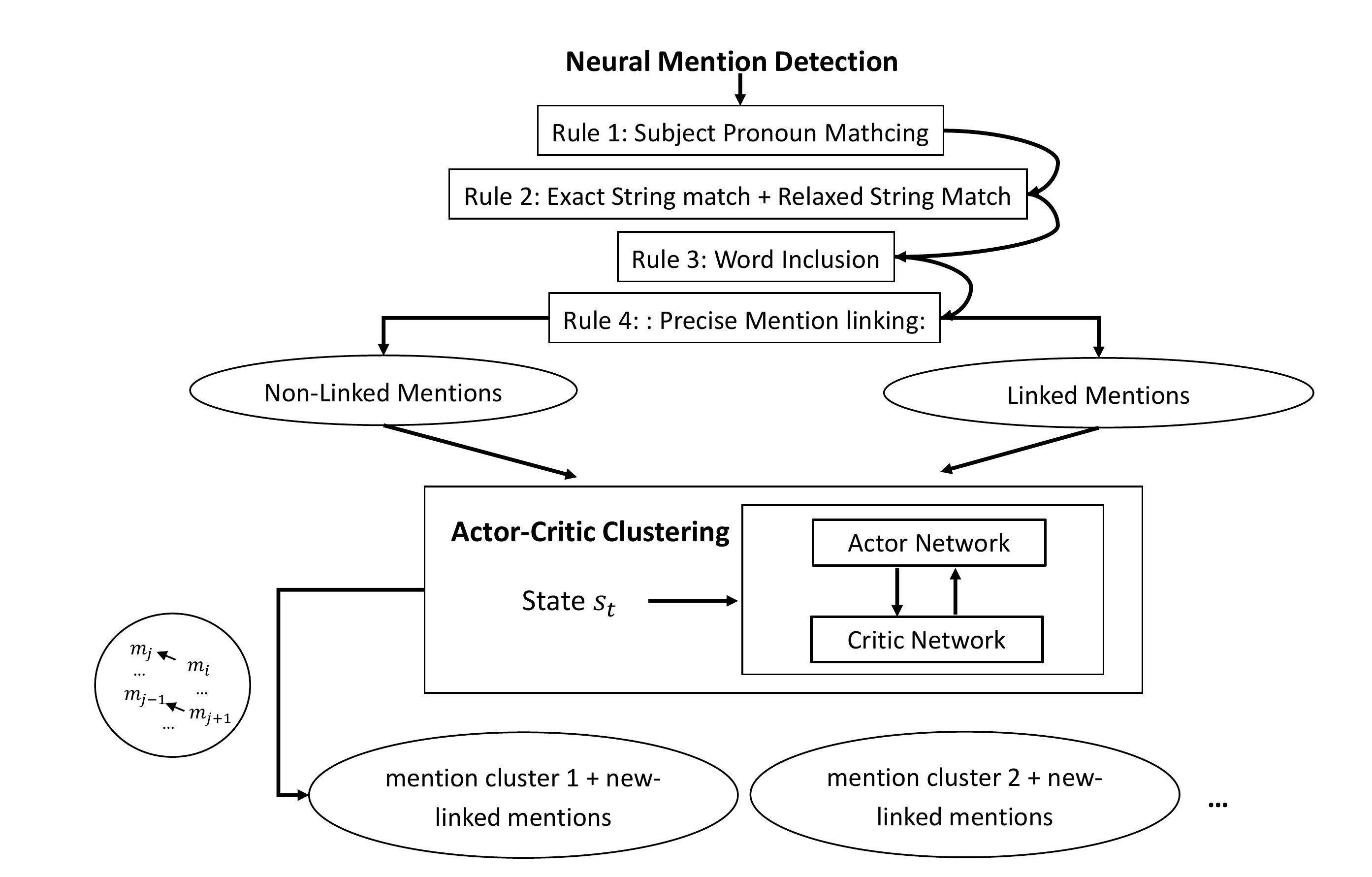}
  \caption{Model structure of HRN-ACC: the model takes the full document as the input to the hybrid rules, and then take two mentions $[m_i,m_j]$ as the actor-critic model's input at each round, where all text spanning up to $K=10$ words is considered.}
  \label{training}
\end{figure*}

In this section, we will give all the design details of the proposed hybrid rule-neural actor-critic coreference (HRN-ACC) model. The entire system structure is proposed as in Figure \ref{training}. As shown the in the figure, the system contains two sub-systems, the first sub-system contains multiple coreference rules and the second part is a actor-critic based coreference clustering model. As discussed in the introduction, both part has their own weakness if working independently, due to their technical limitations. We design the system in such a manner that the systems contains the complementary advantages from both sub-systems; and at the same time, their weaknesses are compensated by each other.

\subsubsection{Coreference Rules}
Inspired by the entity-centric deterministic coreference rules in \cite{lee2013deterministic}, we leverage the following coreference rules to work collaboratively with our proposed actor-critic neural coreference model.\\
{\bf\underline{Rule 1: Subject Pronoun Matching}} The first coreference rule is to match the subject pronouns in a sentence. Here, we use a simple heuristic that searches for the subjects of reporting verbs (\emph{e.g.}, say) in the same sentence. All the first person pronouns {\it{I}}, {\it{my}},{\it{me}} or {\it{mine}} are coreferent. Similarly, it is true for second person pronouns and their plural versions as well. For example, {\it{I}} and {\it{my}} in this sentence are coreferent: ``[I] voted for him because he was most aligned with [my] values."

There are also some sub-rules to work together with rule 1:
\begin{itemize}
\vspace{-0.2cm}\item Nominal mentions cannot be coreferent with \emph{I,you,we} in the same sentence
\vspace{-0.2cm}\item Two different person pronouns in the same sentence cannot be coreferent.
\end{itemize}
{\bf\underline{Rule 2: Exact and Relaxed String Matching}} The second rule is based on textual matching, which can also be summarized as two sub-rules: exact and relaxed string matching rules.
\begin{itemize}
\vspace{-0.2cm}\item Exact string matching: links two mention spans only if they contain exactly the same extent text, including modifiers and determiners.
\vspace{-0.2cm}\item Relaxed string matching: two nominal mentions are coreferent if the strings obtained by dropping the test following their head words as identical. One example is [Clinton] v.s. [Clinton, who was the US president].
\end{itemize}
{\bf\underline{Rule 3: Word Inclusion}} The third conference rule contains two criteria, i.e. If
\begin{itemize}
\vspace{-0.2cm}\item the mention head word matches any head word of mentions in the antecedent entity , and 
\vspace{-0.2cm}\item all the non-stop words in the current entity to be solved are included in the set of non-stop words in the antecedent entity,
\end{itemize}
Then we say the the current mention entity/span is coreferent to its antecedent. For example: [Hilton Dollar Tree hotel] v.s. its antecedent [Hilton Hotel].\\
{\bf\underline{Rule 4: Precise Mention Linking}} This rule links two mentions if any of the following precise mention linking conditions are satisfied:
\begin{itemize}
\vspace{-0.2cm}\item Predicate nominative: the two mentions (nominal or pronominal) are in a copulative subject$–$object relation, for example: [The cheapest flight to NY tomorrow]$_1$ is [the 4:30pm flight by American Airline]$_2$. These two mentions can be linked to each other

\vspace{-0.2cm}\item Relative Pronoun: the mention is a relative pronoun that modifies the head of the antecedent noun, for instance: 
[the hotel [which$]_1$ is closest to airport$]_2$

\end{itemize}

{\bf{Remarks:}} Due to the space limitation, we cannot give all the detailed reasons for each of the coreference rules given in this section. Readers who are interested about the details can refer to \cite{lee2013deterministic} for more detailed explanations.

As shown in Figure \ref{training}, after the coreference rule step, all detected mentions are classified as two groups: linked mentions group $\mathcal{G_L}$ and non-linked mentions group $\mathcal{G_N}$. Within the linked mentions group, there are many mention clusters and each cluster contains all linked mentions presenting the same entity. The next step is to use the actor-critic coreference clustering model to further pick out the missed mention entities from the non-linked mention group, and try to link them with one of the mentions in the linked mention group, such that we won't miss any coreference links that cannot covered by the coreference rules given in this section.
\subsubsection{Actor-Critic Coreference Clustering Model}
As shown in Figure \ref{training}, the actor-critic model is trained to generate the link between a mention $m_i$ to one of its antecedent mention $m_j$.

In general, an actor-critic model is trained based on deep reinforcement learning (DRL) technique. Deep reinforcement learning has been widely used in a variety of NLP and machine learning tasks \cite{wang2018deep,wang2018boosting,wang2019deep,narendra2016fast,wang2019neural,wang2020actor,wang2021coarse}
The model input is a state $s_t$, and its output is an action $a_t$. The model is trained with the target to maximize the expected reward generated by the reward function $r_t$. Specifically, their definitions in our system are as given below:

{\bf{State:}}
The actor-critic model's state $s_t$ is defined by concatenating the current mention span representation $m_i$ with one of its previous spans $m_j$, \emph{i.e.}  $s_t = [m_i,m_j]$.The state contains information of both spans to decide whether or not $m_j$ is an antecedent of $m_i$. It is also worth noticing that both $m_i$ and $m_j$ can be from either the non-linked mention group or the linked mention group. We will later classify them into the same group if $m_j$ is an antecedent of $m_i$ and either one of them has been allocated in the linked group $\mathcal{G_L}$ by the previous coreference rules.

\begin{figure*}[ht]
  \centering
  \includegraphics[width=0.9\linewidth]{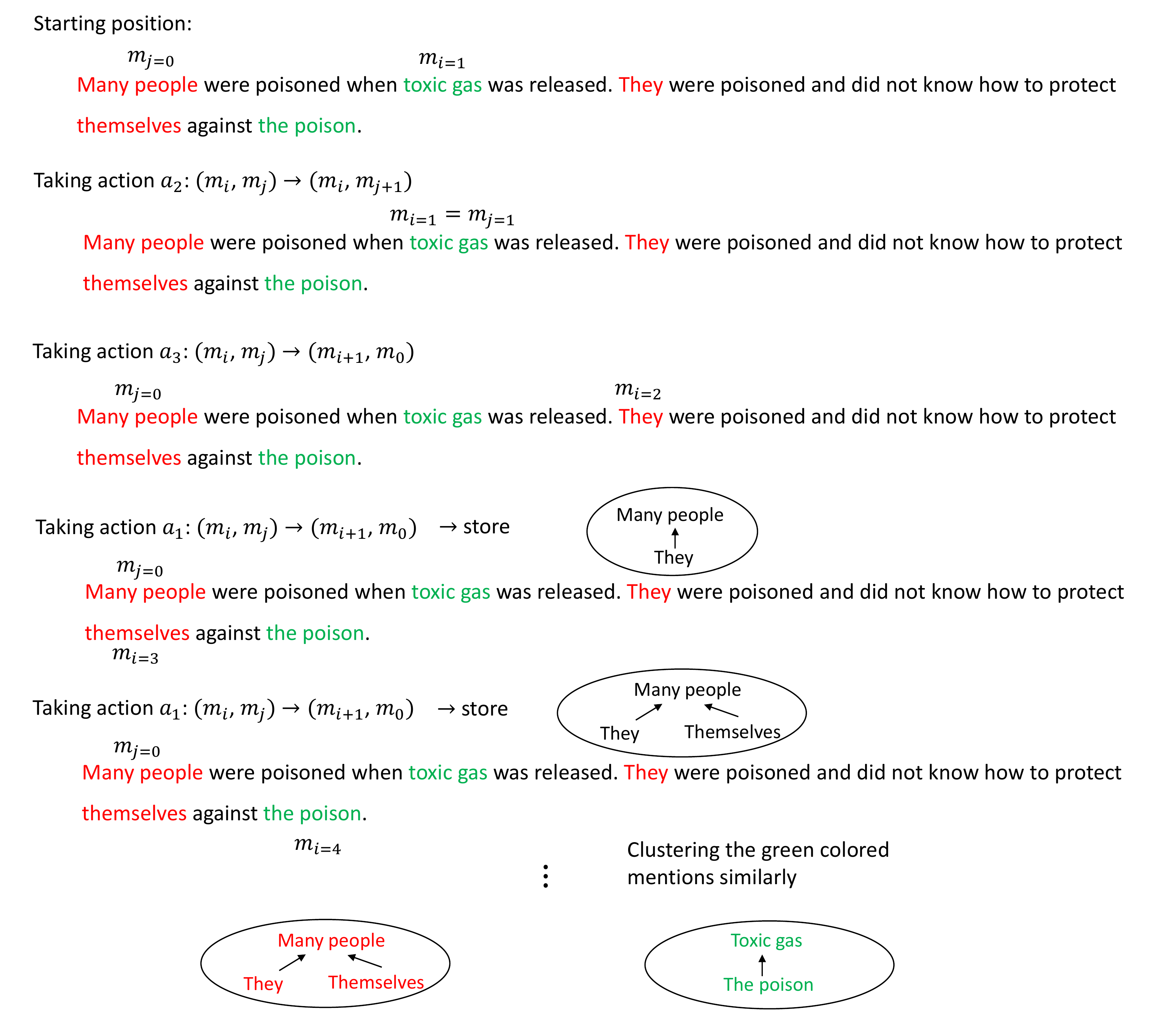}
  \caption{An example using different actions to achieve mention clustering and coreference}
  \label{fig:MentionTransition}
\end{figure*}

{\bf{Action:}}
There are three different types of actions defined in this actor-critic system. The first one is to move the current span $m_i$ by one step to the right, \emph{i.e.} $m_i \rightarrow m_{i+1}$, which indicates that $m_i$ has found an antecedent $m_j$, and hence moves to the next span. At the same time, the second term representing the antecedent span is reset to $m_0$, and the state $s_t=[m_i,m_j$] is stored as $m_j$, which is an antecedent of $m_i$ formed by linking the two mentions. On the other hand, if the previous span $m_j$ is not an antecedent of $m_i$, the system should move to $m_j$'s next span $m_{j+1}$ for further evaluation. There is a third action: if input $s_t=[m_i$,$m_j$] satisfies the condition that $i=j+1$, \emph{i.e.} $m_i$ is the next mention of $m_j$ in the context, it is possible that there is no antecedent for $m_i$, and action $a_t$ is to transfer the system from current state $s_t=[m_i,m_j]$ to the new state $s_{t+1}=[m_{i+1},m_0]$, but the antecedent link is not stored as action 1. To represent these three actions mathematically, we have the following:

\begin{equation}
a_t= \begin{cases}
               [m_i,m_j]\rightarrow[m_{i+1},m_0],\text{store }[m_i,m_j]\\
               [m_i,m_j]\rightarrow[m_{i},m_{j+1}]\\
               [m_i,m_j]\rightarrow[m_{i+1},m_{0}],\text{don't store}
            \end{cases}
\label{action}
\end{equation} 
Once an action $a_t$ state $s_t$ will then transition to $s_{t+1}$, directed by action $a_t$. A graphical illustration on how the actions perform mention clustering is given in Figure \ref{fig:MentionTransition}.

As shown in Figure \ref{fig:MentionTransition}, the system takes action $a_2$ at the first step by moving $m_j$ to the right by one step. Since the system finds that $m_j$ is not an antecedent of $m_i$, then it takes action $a_3$ to reset $j=0$ and starts the mention search again. In the next two steps, the system sequentially chooses action $a_2$ and $a_1$ , hence it will store the tuple $[m_i,m_j]$ which are given at the system step taking action $a_2$. The action selection process repeats until both $m_i$ and $m_j$ reach the last mention of the input sample.

{\bf{Reward:}}
Inspired by \cite{dozat2016deep} and \cite{zhang2018neural}, the reward $r_t$ at a state $s_t$ is defined using a biaffine attention technique to model the likelihood of $m_j$ as an antecedent of $m_i$, by jointly considering the distance between two mentions. The reward is hence defined as:
\begin{equation}
\begin{split}
r_t &= e^{-\gamma \parallel i-j \parallel}(v_m^Tf_{nn}^1(m_i)+v_m^Tf_{nn}^1(m_j)\\
&+f_{nn}^2(m_i,m_j,m_i\circ m_j)+v_{bi}^Tf_{nn}^3(m_i))
\end{split}
\label{reward}
\end{equation} 
where $f_{nn}^{1,2,3}(\cdot)$ are feedforward networks to reduce span representation dimensions, ${v}_{bi} \in \mathbb{R}^{k\times 1}$ is the linear transformation matrix/vector, and $k$ is the dimension of the output of $f_{nn}(\cdot)$, $\gamma \in (0,1)$ is the user chosen decay factor, $\circ$ denotes element-wise multiplication. The exponential decay term $e^{-\gamma \parallel i-j \parallel}$ indicates that a smaller reward is assigned when the distance between two mentions ($m_i$,$m_j$) (\emph{i.e.} $\parallel i-j \parallel$) becomes larger. The reward design based on two mentions' distance follows a straight-forward observation that: the probability of two mentions are coreferent to each other becomes smaller when distance between them becomes larger.

As the reward defined in \ref{reward}, the terms $v_m^Tf_{nn}^1(m_i)$ and $v_m^Tf_{nn}^1(m_j)$ are the mention scores to measure their likelihoods as entity mentions. $f_{nn}^2(m_i,m_j,m_i\circ m_j)$ defines the compatibility between $m_i$ and $m_j$, and ${v}_{bi}^Tf_{nn}^3(m_i)$ measures the likelihood of the current mention $m_i$ having an antecedent. The target of the actor-critic RL system is to maximize the expected rewards given that $m_j$ is an antecedent of $m_i$. 

It is worth noticing that $f_{nn}^1$, $f_{nn}^2$ and $f_{nn}^3$ are all feedforward neural networks with unknown weights need to be learnt. Hence we follow the below label definitions to train these networks concurrently with our actor-critic system:

\begin{equation}
\begin{cases}
\begin{split}
f_{nn}^{1}(m_i)&=1 \;\;\; \text{if } m_i \text{ is an entity mention}\\
f_{nn}^{1}(m_i)&=0 \;\;\; \text{if } m_i \text{ is not an entity mention}\\
f_{nn}^{2}(\cdot)&=1 \;\;\; \text{if }m_j\text{ is an antecedent of } m_i\\
f_{nn}^{2}(\cdot)&=0 \;\;\; \text{if }m_j\text{ is not antecedent of } m_i\\
f_{nn}^{3}(m_{i})&=1 \;\;\;\text{if } m_i \text{ has an antecedent} \\
f_{nn}^{3}(m_{i})&=0 \;\;\;\text{if } m_i \text{ has no antecedent} 
\end{split}
\end{cases}
\end{equation}

During each iteration of actor-critic model training, we will train the three neural networks $f_{nn}^{1,2,3}$ with the given batch sample(s) first and then train the actor-critic models using the reward defined by the outputs of $f_{nn}^{1,2,3}$.

\subsubsection{Re-union Entity Mentions}
After the actor-critic coreference clustering step, the system generates many mention clusters, where each cluster contains several mention spans referring to the same entity. For a given cluster $\mathcal{C_N}$, if any one of the entities, say $m_k \in \mathcal{C_N} \subset \mathcal{G_N}$, is also in another cluster $\mathcal{C_L}$ in the linked mention group $\mathcal{G_L}$, $m_k \in \mathcal{C_L} \subset \mathcal{G_L}$, our system will union the cluster $\mathcal{C_N}$ with the cluster in the linked mention group containing $m_k$, \emph{$\mathcal{C_L}$}. To represent it mathematically, the new re-union entity cluster $\mathcal{C_U}$ is:
\begin{equation}
\mathcal{C_U}=\mathcal{C_N}\cup \mathcal{C_L}
\end{equation}

\section{Model Training}
\subsection{Actor-Critic Model Training}
As mentioned previously, in this hybrid system, we use an actor-critic neural coreference model, where two neural networks are trained to model the actor and critic separately. One important reason that we use an actor-critic model instead of other reinforcement learning models is because it is known to converge more smoothly and have better training performance on a system with a large state space \cite{lillicrap2015continuous}, which is like our scenario.

%

There are also two loss functions corresponding to the actor and critic networks, specified as:
\begin{equation}
\begin{split}
\mathcal{L}_{actor}&=-\log \pi_{\theta} (a_t|s_t)(r_t+\gamma V(s_{t+1})-V(s_t))\\
\mathcal{L}_{critic}&=(r_t+\gamma V(s_{t+1})-V(s_t))^2
\end{split}
\end{equation}
where $\gamma$ is a discount factor, and $\pi_{\theta}$ is the policy probability of taking action $a_t$. 



\subsection{Joint Training with Mention Detection}
Our actor-critic model performs mention clustering after two steps, \emph{i.e.} mention detection and coreference rules. As mentioned before, our mention detection model relies on a feed-forward neural network mention classifier $f_{nn}^1(m_i)$ to generate the mention score to measure its likelihood as an entity's mention. During training, only mention cluster labels are available, rather than antecedent links, and hence we propose two augmented joint loss functions for training our actor-critic model by also taking the mention detection loss into consideration. Before that, we must first define the mention detection loss as:
\begin{equation}
\begin{split}
\mathcal{L}_{detect}&= y(m_i)\log(S(m_i))\\
&+(1-y(m_i))\log(1-S(m_i))
\end{split}
\end{equation}
where $y(m_i)$ is equal to 1 if $m_i$ is in one of the gold mention clusters, and otherwise is equal to 0. $S(m_i)$ is a sigmoid function of $m_i$.

Since mention detection is a prerequisite for both actor and critic network, both of their loss functions are augmented by the detection loss $\mathcal{L}_{detect}$ as:
\begin{equation}
\begin{split}
\mathcal{L}_{actor}^{\prime}&=\mathcal{L}_{actor}+\mathcal{L}_{detect}\\
\mathcal{L}_{critic}^{\prime}&=\mathcal{L}_{critic}+\mathcal{L}_{detect}
\end{split}
\label{lossac}
\end{equation}

{\it{Remarks:}} Mention detection can also treated as a separate semantic parsing task, and can be achieved by generic tagging and parsing models for natural language understanding as in \cite{wang2018bi,wang2018new,wang2018neural,wang2019deep,wang2020interactive,wang2020bi,wang2020new,wang2020multi,shen2021system,wang2021adversarial}. 
\begin{table*}[t]\scriptsize
\parbox{1\linewidth}{
\centering
	\caption{Experimental results on CoNLL-2012 English test set}
	\label{table:comparisonCoNLL}
	\begin{tabular}{>{\centering\arraybackslash}p{4cm}|>{\centering\arraybackslash}p{0.5cm}>{\centering\arraybackslash}p{0.5cm}>{\centering\arraybackslash}p{0.5cm}>{\centering\arraybackslash}p{0.4cm}>{\centering\arraybackslash}p{0.5cm}>{\centering\arraybackslash}p{0.5cm}>{\centering\arraybackslash}p{0.5cm}>{\centering\arraybackslash}p{0.5cm}>{\centering\arraybackslash}p{0.5cm}>{\centering\arraybackslash}p{0.6cm}}
		\toprule
		\multirow{1}{*}{\textbf{}} & \multirow{1}{*}{\makecell{\textbf{}}} & \multirow{1}{*}{\makecell{\textbf{$MUC$}}}& \multirow{1}{*}{\makecell{\textbf{}}}& \multirow{1}{*}{\makecell{\textbf{}}} & \multirow{1}{*}{\makecell{\textbf{$B^3$}}}& \multirow{1}{*}{\makecell{\textbf{}}}& \multirow{1}{*}{\makecell{\textbf{}}} & \multirow{1}{*}{\makecell{\textbf{$CEAF_{\phi_4}$}}}& \multirow{1}{*}{\makecell{\textbf{}}}& \multirow{1}{*}{\makecell{\textbf{}}}\\		
		\midrule
		\multirow{1}{*}{\textbf{Model}} & \multirow{1}{*}{\makecell{\textbf{P}}} & \multirow{1}{*}{\makecell{\textbf{R}}}& \multirow{1}{*}{\makecell{\textbf{F1}}}& \multirow{1}{*}{\makecell{\textbf{P}}} & \multirow{1}{*}{\makecell{\textbf{R}}}& \multirow{1}{*}{\makecell{\textbf{F1}}}& \multirow{1}{*}{\makecell{\textbf{P}}} & \multirow{1}{*}{\makecell{\textbf{R}}}& \multirow{1}{*}{\makecell{\textbf{F1}}}& \multirow{1}{*}{\makecell{\textbf{Avg. F1}}}\\
\midrule 
		\multirow{2}{*} {}{\textbf{HRN-ACC}}  & 94.7 &90.8 &93.5 &88.5 &85.9&87.4 &85.6 &83.3 &85.4 &{\bf{88.6}}\\
		\midrule
		
		\multirow{2}{*} {}\cite{wu2020corefqa}  & 88.6 &87.4 &88.0 &82.4 &82.0 &82.2 &79.9 &78.3 &79.1 &83.1\\
		\multirow{2}{*} {}\cite{joshi2020spanbert}  & 85.8 &84.8 &85.3 &78.3 &77.9 &78.1 &76.4 &74.2 &76.3 &79.6\\
		\multirow{2}{*} {}\cite{joshi2019bert}  & 84.7 &82.4 &83.5 &76.5 &74.0 &75.3 &74.1 &69.8 &71.9 &76.9\\
				
		\multirow{2}{*} {}\cite{kantor2019coreference}  & 82.6 &84.1 &83.4 &73.3 &76.2 &74.7 &72.4 &71.1 &71.8 &76.6\\ 
		
		\multirow{2}{*} {}\cite{fei2019end}  & 85.4 &77.9 &81.4 &77.9 &66.4 &71.7 &70.6 &66.3 &68.4 &73.8\\ 
				
		\multirow{2}{*} {}\cite{lee2018higher}  & 81.4 &79.5 &80.4 &72.2 &69.5 &70.8 &68.2 &67.1 &67.6 &73.0\\ 
		
		\multirow{2}{*} {}\cite{zhang2018neural}  & 79.4 &73.8 &76.5 &69.0 &62.3 &65.5 &64.9 &58.3 &61.4 &67.8\\ 
		
		\multirow{2}{*}{}\cite{lee2017end}  &78.4 &73.4 &75.8 &68.6 &61.8 &65.0 &62.7 &59.0 &60.8 & 67.2  \\
		
		\multirow{2}{*}{}\cite{clark2016deep}  &79.2 &70.4 &74.6 &69.6 &58.0 &63.4 &63.5 &55.5 &59.2 & 65.7  \\
		\multirow{2}{*}{}\cite{wiseman2016learning}  &77.5 &69.8 &73.4 &66.8 &57.0 &61.5 &62.1 &53.9 &57.7 & 64.2  \\
		\bottomrule		
	\end{tabular}
}
\end{table*}

\begin{table}[h]\scriptsize
\parbox{1\linewidth}{
\centering
	\caption{Ablation study between the HRN-ACC model and a supervisely trained BERT-based neural coreference model}
	\label{table:ablationstudy2}
	\begin{tabular}{>{\centering\arraybackslash}p{5cm}|>{\centering\arraybackslash}p{1cm}}
		\toprule
		\multirow{1}{*}{\textbf{Model}} & \multirow{1}{*}{\makecell{\textbf{Avg. F1}}}\\
		\midrule	
		\multirow{2}{*}{}HRN-ACC Model & 88.6\\ 	
		\multirow{2}{*}{}Supervised LSTM w BERT &79.9\\
		\midrule		
		\multirow{2}{*}{}\cite{wu2020corefqa}  &83.1\\
		\bottomrule		
	\end{tabular}
}
\end{table}

\begin{table}[h]\scriptsize
\parbox{1\linewidth}{
\centering
	\caption{Ablation study on models with/without mention detection loss}
	\label{table:ablationstudy}
	\begin{tabular}{>{\centering\arraybackslash}p{5cm}|>{\centering\arraybackslash}p{1cm}}
		\toprule
		\multirow{1}{*}{\textbf{Model}} & \multirow{1}{*}{\makecell{\textbf{Avg. F1}}}\\
		\midrule
		\multirow{2}{*}{}HRN-ACC & 88.6\\ 	
		\multirow{2}{*}{}HRN-ACC without mention detection  &86.2\\
		\midrule		
		\multirow{2}{*}{}\cite{wu2020corefqa}  &83.1\\
		\bottomrule		
	\end{tabular}
}
\end{table}

\section{Experiment}
\subsection{Dataset}
The dataset we used for experimentation is the CoNLL-2012 Shared Task English data \cite{pradhan2012conll} based on the OntonNotes corpus. The training set contains 2,802 documents, and the validation set and test set contain 343 and 348 documents, respectively. Similar to the previous works, we use three different metrics: MUC \cite{vilain1995model}, B$^3$ \cite{bagga1998algorithms} and CEAF$_{\phi_4}$ \cite{luo2005coreference}, and report the respective precision, recall and F1 scores.

\subsection{Model Implementation}

In this paper, we consider all spans up to 250 antecedents and 10 words.
In our BERT setup, we use the pre-trained BERT embeddings giving in \cite{devlin2018bert}. Both the actor and critic networks use the LSTM structure with hidden layer size of 200. The feedforward neural networks used to generate mention detection scores include two hidden layers with 150 units and ReLU activations. To be comparable with previous popular models, we include features (speaker ID, document genre, span distance and span width) as the 20-dimensional learned embeddings. For the $f_{nn}^{(1,2,3)}$, we use a dropout rate of 0.5.
 
\begin{table}[h]\scriptsize
\parbox{1\linewidth}{
\centering
	\caption{Ablation study on models with/without coreference rules}
	\label{table:ablationstudycoref}
	\begin{tabular}{>{\centering\arraybackslash}p{3cm}|>{\centering\arraybackslash}p{1cm}>{\centering\arraybackslash}p{1cm}>{\centering\arraybackslash}p{1cm}}
		\toprule
		\multirow{1}{*}{\textbf{Model}} & \multirow{1}{*}{\makecell{\textbf{Avg. P}}}& \multirow{1}{*}{\makecell{\textbf{Avg. R}}}& \multirow{1}{*}{\makecell{\textbf{Avg. F1}}}\\
		\midrule
		\multirow{2}{*}{}HRN-ACC &90.5 &86.8 & 88.6\\ 	
		\multirow{2}{*}{}HRN-ACC without coreference rules  &85.8& 85.2 &85.5\\
		\multirow{3}{*}{}Heuristic coreference rules &78.2 &66.9 &72.1\\
		\midrule		
		\multirow{2}{*}{}\cite{wu2020corefqa} &85.3 &81.0 &83.1\\
		\bottomrule		
	\end{tabular}
}
\end{table}

\begin{table*}[h]\scriptsize
\parbox{1\linewidth}{
\centering
	\caption{Mention detection subtask on development set}
	\label{table:mentiondetection}
	\begin{tabular}{>{\centering\arraybackslash}p{5cm}|>{\centering\arraybackslash}p{2cm}>{\centering\arraybackslash}p{2cm}>{\centering\arraybackslash}p{2cm}>{\centering\arraybackslash}p{2cm}}
		\toprule
		\multirow{1}{*}{\textbf{Model/Span Width}} & \multirow{1}{*}{\makecell{\textbf{ 1-2}}}& \multirow{1}{*}{\makecell{\textbf{ 3-4}}}& \multirow{1}{*}{\makecell{\textbf{5-7}}}& \multirow{1}{*}{\makecell{\textbf{8-10}}}\\
		\midrule
		\multirow{2}{*}{}HRN-ACC & 93.6& 84.2& 73.6& 66.2\\ 	
		\midrule		
		\multirow{2}{*}{}\cite{joshi2020spanbert}  &87.6& 76.4& 63.4& 52.8\\
		\bottomrule		
	\end{tabular}
}
\end{table*}
\subsection{Performance}
In Table \ref{table:comparisonCoNLL}, we compare our model with the previous state-of-the-art models. Here, we only compare single models without any ensemble for fairness. It can be observed that the HRN-ACC model with BERT pre-trained embeddings achieves the best performance in terms of all metrics. Especially compared to the model using a similar biaffine technique, but being trained in a supervised manner instead (\emph{i.e.} using given labels for training instead of training the model through maximizing the expected rewards), HRN-ACC model generates a higher recall value since both the heuristic coreference rules sub-module and the actor-critic training technique can cover more stochastic mention patterns or antecedent links in test test, which are not exhibited in the training data.

\subsection{Ablation Study}
\subsubsection{Effect of Coreference Rules}
In order to evaluate the impact of corefernce rules in our system, we compare our HRN-ACC model with two other setups: one model is without heuristic rules by feeding all mentions into the actor-critic model, and then do clustering based on the actor-critic model only. The other model only retains the four coreference rules and use them to perform coreference resolution. The results are given in Table \ref{table:ablationstudycoref}.

Based on the results in the table, we can see that all metrics drop after we remove the coreference rules, especially the average precision drops by 4.7$\%$. This is mainly because the rules are stricter and more accurate despite their limited coverage. This also explains why only using rules gives a huge recall drop by 19.9$\%$ since there are lots of coreference links cannot be covered by the rules only.

\subsubsection{Effect of Joint Training with Mention Detection}
To evaluate the impact of the mention detection upon the mention cluster task, we remove the mention detection term in the loss function given in \ref{lossac}. The result is given in Table \ref{table:ablationstudy}. We can observe that the overall F1 score decreases by 2.4 on HRN-ACC model. The drop is mainly because the model does not leverage any useful information from mention detection in this setup.
\subsubsection{Effect of Actor-Critic Reinforcement Learning Model}
In order to evaluate the impact of our semi-supervisely trained actor-crtic reinforcement learning model over the task, we replace our actor-critic reinforcement learning model structure by a supervised BERT model with the same structure as the actor network, \emph{i.e.} a LSTM structure with a hidden layer size of 200. The feedforward neural networks used to generate mention detection scores are kept as the same network with 150 units and ReLU activations. In order to train the model in a supervised manner, we use the reward $r_t$ as the coreference score or the training label for each given mention pairs fed into the system. The inputs of the training system are the mention representations of a document sample as in \cite{zhang2018neural}. The mentions scores and loss functions are kept as the same as in our actor-critic setup. During inference, mentions are clustered by using the generated coreference score in a given sample.

The results given in Table \ref{table:ablationstudy2} shows that the model trained using actor-critic network structure performs far better than the same model trained supervisely without using the DRL structure. One main reason is that the supervisly trained model is not able to capture the mentions' stochasticity, especially when the same mentions appear in different contexts.

{\bf{Remarks:}} Here we define the reinforcement learning as a semi-supervised training technique as it doesn't use the label as ground truth, but with a target to maximize the reward function containing the label value. Comparatively, most of the deep learning models are supervised models as their loss functions are defined using labels directly.

\subsection{Mention Detection Task}
To further understand our model, we separate the mention detection task from the joint task. In this setup, we consider spans with mention scores higher than zero as mentions. The mention detection accuracy versus span length is as shown in Table \ref{table:mentiondetection}. Our model with different embeddings both perform better than the current state-of-the-art model on the CoNLL-2012 dataset in terms of mention detection accuracy. It is also observed that the advantage becomes larger when the span width increases.

\section{Related Works}
There has been a long history of using machine learning algorithms to tackle the coreference resolution tasks \cite{ng2010supervised}, most of these systems are either entity-level models \cite{wiseman2016learning,clark2016improving} or mention-ranking models \cite{lee2017end,wiseman2015learning,clark2016deep} These models are typically trained with heuristic loss functions that assign costs to different error types. Many systems also incorporate entity-level information using joint inference \cite{haghighi2010coreference}and systems that build up coreference clusters incrementally\cite{raghunathan2010multi}. Most of these models use syntactic parsers for head features and contain hand engineered mention proposal algorithms, and many of them leverages distributed representations of mention pairs or mention clusters to reduce the number of hand-crafted features. 

\cite{lee2017end} proposed an end-to-end coreference resolution model by directly considering all spans in a document as potential mentions and learn distributions over possible antecedents for each. The introduced model computes span embeddings that combine context-dependent boundary representation with a head-find attention mechanism. Their span-ranking approach is similar to mention ranking algorithms, but reasoning over a larger space by jointly detecting mentions and predicting coreference. In our paper, we also refer to a similar span representation structure as given in this work.

Another end-to-end coreference resolution model given by \cite{zhang2018neural} leverages the biaffine attention model instead of pure feed forward networks to computer antecedent scores, and the model is being trained by jointly optimizing the mention detection accuracy and the mention clustering log-likelihood given the mention cluster labels. In our system, we also refer their biaffine model structure to improve our actor-critic model performance.

Recently, there are many models leveraging BERT structure \cite{devlin2019bert,joshi2019bert,kantor2019coreference,joshi2020spanbert} and other unsupervised contextualized representations which generate impressive gains on coreference resolution tasks. Among them, some use BERT pretraining on passage-level sequences to more effectively model long-range dependencies \cite{kantor2019coreference,joshi2019bert}, some other models also adopt another SpanBERT representation \cite{joshi2020spanbert} focusing on pretraining span representations which also achieves outstanding performance on several benchmark datasets.

\section{Conclusion}
In this paper, we propose a hybrid rule-neural coreference resolution system based on actor-critic learning. The system is built in a way such that it can leverage the advantages from both the heuristic rules and a actor- critic neural conference model. This end-to-end system can also perform both mention detection and resolution by leveraging a joint training algorithm. Our model with the BERT span representation achieves the state-of-the-art performance among the models on the CoNLL-2012 Shared Task English Test Set.

\bibliography{acncr}
\bibliographystyle{acl_natbib}

\end{document}